\documentclass[letterpaper]{article}
\usepackage{aaai}
\usepackage{times}
\usepackage{helvet}
\usepackage{courier}
\usepackage{graphicx}
\usepackage{xcolor}
\graphicspath{ {./} }
\newlength{\Oldarrayrulewidth}
\newcommand{\Cline}[2]{%
  \noalign{\global\setlength{\Oldarrayrulewidth}{\arrayrulewidth}}%
  \noalign{\global\setlength{\arrayrulewidth}{#1}}\cline{#2}%
  \noalign{\global\setlength{\arrayrulewidth}{\Oldarrayrulewidth}}}

\begin{document}
%

\title{Breaking Free Transformer Models: Task-specific Context Attribution Promises Improved Generalizability Without Fine-tuning Pre-trained LLMs}

\author{Stepan Tytarenko,\\
{Fordham University, New York}\\
stytarenko@fordham.edu
\And
Dr. Mohammad Ruhul Amin,\\
{Fordham University, New York}\\
mamin17@fordham.edu}

\maketitle
\begin{abstract}
\begin{quote}
Fine-tuning large pre-trained language models (LLMs) on particular datasets is a commonly employed strategy in Natural Language Processing (NLP) classification tasks. However, this approach usually results in a loss of models’ \textit{generalizability}. 
In this paper, we present a framework that allows for maintaining \textit{generalizability}, and enhances the performance on the downstream task by utilizing task-specific context attribution.  
We show that a linear transformation of the text representation from any transformer model using the task-specific concept operator results in a projection onto the latent concept space, referred to as context attribution in this paper. The specific concept operator is optimized during the supervised learning stage via novel loss functions. 
The proposed framework demonstrates that context attribution of the text representation for each task objective can improve the capacity of the discriminator function and thus achieve better performance for the classification task. 
Experimental results on three datasets, namely HateXplain, IMDB reviews, and Social Media Attributions, illustrate that the proposed model attains superior accuracy and \textit{generalizability}. 
Specifically, for the non-fine-tuned BERT on the HateXplain dataset, we observe \textbf{8\%} improvement in accuracy and \textbf{10\%} improvement in F1-score. Whereas for the IMDB dataset, fine-tuned state-of-the-art XLNet is outperformed by \textbf{1\%} for both accuracy and F1-score. 
Furthermore, in an out-of-domain cross-dataset test, DistilBERT fine-tuned on the IMDB dataset in conjunction with the proposed model improves the F1-score on the HateXplain dataset by \textbf{7\%}. 
For the Social Media Attributions dataset of YouTube comments, we observe \textbf{5.2\%} increase in F1-metric. The proposed framework is implemented with PyTorch and provided open-source on GitHub\footnote{https://github.com/StepanTita/space-model}. 
\end{quote}
\end{abstract}

\section{Introduction}
Currently, the domain of Language Models is one of the most rapidly developing areas of machine learning. Transformer architecture \cite{NIPS2017_3f5ee243} has proven itself as a state-of-the-art approach towards the absolute majority of Natural Language Processing (NLP) domains \cite{9364465}. A particular strength of the language models is their \textit{generalizability} \cite{swamy-etal-2019-studying}, \cite{9671970}. 
\\
By pre-training the model on a big chunk of semi-structured data and then fine-tuning with task-specific labeled data, we may obtain state-of-the-art performance in the problems of classification, regression, language translation, and more. However, the important note here is that the most crucial pre-training stage is usually costly, and repeating it for every new task is computationally inefficient.
At the same time, the fine-tuning only downstream task specific head of pre-trained models is time efficient and requires much less labeled data, preserving models’ \textit{generalizability}. On the other hand, this part of the pipeline might be a bottleneck to the process. Usually, single fine-tuning does not produce results on par with the complete model adaptation via training \cite{peters-etal-2019-tune}. Explaining and adapting the results to the various downstream tasks is also tricky. To avoid any confusion in this paper, we are going to use term "training" or "model adaptation" referring to the process of full model retraining (adapting all of the weights), while for the head-only adaptation we are going to use term "fine-tuning". Fine-tuning is usually easier, faster and requires less data.

\bigskip

As a potential solution to the problem, we propose a novel method of model fine-tuning. We call this approach the \textbf{Space Model}. The whole idea is to replace the classification head of the transformer model with a set of conceptual operators, projecting the contextual embeddings of the model to the set of concept spaces referred to as context attributions. The Space model is an additional model framework that plays the role of the pipeline’s original downstream task head. In this work, we limit ourselves to the review of the classification capabilities of the proposed approach, but generally, this is not a limitation to the technique in any way.

\bigskip

The model is designed in a way that a set of concepts describes different classes; such a set is called the “context attribution”. It is worth noting that we do not limit these concepts in terms of overlapping. Some context attributions might overlap if that makes sense in terms of the problem solved. This paper reviews one such task where overlapping context attributions are entirely appropriate. What we would like to avoid is allowing multiple concepts to converge to the same representation. For that type of regularization, we introduce an additional loss called \textbf{Intra-Space} loss. Its goal is to make sure concepts in the context attribution are disjoint.

\bigskip

As was stated previously, the Space Model is an external framework with a set of operators on top of the transformer model. Generally speaking, this is not limited to the transformer architecture either. Potentially, any technique that can produce embeddings may be used as the Space model’s base model, such as Word2Vec \cite{word2vec}, Glove \cite{pennington-etal-2014-glove}, or RNN \cite{10.5555/104279.104293}, \cite{Ghosh2016ContextualL}. Further in this paper, whenever we refer to the base model, we mean the model that produces the embeddings for the space model. Some of the base models tested in this paper include BERT \cite{devlin-etal-2019-bert}, DistilBERT \cite{Sanh2019DistilBERTAD}, and XLNet \cite{NEURIPS2019_dc6a7e65}.

\bigskip

The benchmarking and evaluation of the proposed solution are done with various configurations of the base models and across multiple datasets. We test performance for the specific task, fine-tuning the Space model for that particular downstream task, and we also test the performance of the model on the task that is related to the original semantically; however, it uses different data. The baseline for the comparison is mainly the performance of the original base model fine-tuned for the downstream task. During the experiments, we prove that besides an evident performance boost, the Space model also stabilizes the training process and generalizes better for the semantically close tasks. We also prove that the space model can achieve a significant performance boost even when using a smaller number of parameters than the base model fine-tuned on the downstream task.

\bigskip

The datasets used for benchmarking are HateXplain \cite{mathew2021hatexplain}, IMDB reviews sentiment dataset \cite{maas-EtAl:2011:ACL-HLT2011}, and Social Media Attributions dataset of YouTube comments, related to Chennai water crisis \cite{sarkar-etal-2020-social}. The main reason for choosing corresponding datasets is that HateXplain is considered a very complex dataset, with imbalanced data, and labels “offensive” and “hateful” are conceptually very close. On the other hand, IMDB sentiment reviews are a semantically close dataset to the former one and are reasonably easily interpretable. Such a relation is essential since we would like to test the \textit{generalizability} of the proposed technique. Besides, in the Social Media Attributions paper, the authors apply a very similar approach to the one proposed in this paper, however, with additional manual labeling of the concepts and multiple runs. We would like to show that our approach achieves superior performance without additional manual labeling and via a single pass.

The impact and novelty of this paper include:
\begin{itemize}
    \item A novel framework for Language Model fine-tuning, which outperforms the baseline score of the base models such as BERT, DistilBERT, and XLNet
        \begin{itemize}
            \item For the non-pretrained BERT (only trained context attribution operator) we observe \textbf{8\%} improvement in accuracy and \textbf{10\%} in F1-score on HateXplain data
            \item For the IMDB with XLNet base model, we observe an improvement of around \textbf{1\%} after full model adaptation compared to fully trained vanilla XLNet
            \item Space-model with the base model as DistilBERT non-pretrained (only trained context attribution operator) on IMDB dataset, in a zero-shot manner outperforms basic DistilBERT in the same manner (only head fine-tuning) by \textbf{7\%} on F1-score
            \item Compared to the results from the Social Attribution paper (with manual supervision), we observe an improvement of \textbf{5.2\%} with our model without additional supervision
        \end{itemize}
    \item A novel loss function that improves the generalization and stabilization of the training process, improving the zero-shot capacity of the transformers
\end{itemize}

\section{Related work}

The task of effective fine-tuning is one of the main tasks in the modern NLP. Cheaper and faster results of great quality are very appealing and a current trend in the domain. However, we are sourcing the inspiration for our framework not only from the latest NLP findings. The core idea has a root in a Psychological Belief Attribution Theory \cite{BEM19721}, \cite{attribution-theory}. The theory revolves around the idea of attribution of certain concepts with corresponding behavior patterns. The concepts (sometimes also referred to as factors) may be external and internal. These factors are usually related to personal beliefs, and they affect the decisions and behavior of an individual. Researchers have also classified people based on these factors (e.g., pessimistic attribution, optimistic attribution, hostile attribution). We try to apply the same idea to language modeling, attributing certain concepts with class labels. In general, the idea of measuring and researching the belief attribution of language models is not novel. The authors of \cite{hase-etal-2023-methods} have not only proved that certain language models possess the beliefs, but they have also provided metrics to measure such beliefs and a tool to update these beliefs, as well as visualization of beliefs graph.

\bigskip

It is very natural that semi-supervised solutions are mentioned when it comes to fine-tuning with the least resources. These also mainly include ensembling to achieve regularization when working with unsupervised data. One of the first such approaches addressing this issue is the COREG \cite{10.5555/1642293.1642439}. The technique uses two k-nearest-neighbor regressors with different distance metrics to label the data. The distance metric, in that case, would serve as the confidence label. This approach uses a fundamental idea that some features in some spaces are aligned with similar class labels and are further apart from the different class labels. This is an essential fact that is reused in the Space model.

\bigskip

Another later technique involves minimal supervision for the labeling of the concept space, and then, based on this concept space, the model can autonomously label the unlabelled data \cite{inproceedings}. The key idea here is the knowledge extraction from the manually labeled concept space. It is claimed in the work that labeling a set of concepts and then running an algorithm on a set of documents to label them based on these supervised concepts is a superior technique. Our main takeaway from there is that we can extract knowledge from the supervised concept space for unlabelled data. Furthermore, what we would like to propose is testing if this concept space can help us make a prediction at the inference stage rather than during labeling.

\bigskip

Social Media attributions in the Context of Water Crisis paper \cite{sarkar-etal-2020-social} is accomplishing a task very close to the one we are dealing with. However, unlike our approach, same as the previous one, their technique requires supervised sub-concept labeling. Besides, they measure the similarities between sub-concepts and the attention of the Language Model by feeding the sentence to the model multiple times, each time with a new sub-concept. However, they are using the similarity measure to find the concept sub-space that best describes the given sentence and make the decision based on that. Our approach does this all in one pass and in an automated manner. We do not require manual labeling of the concept sub-spaces. We expect to learn them during the fine-tuning phase.

\bigskip

In the paper on Interacting Conceptual Spaces \cite{Bolt2019}, the authors create all of the necessary mathematical background required to formulate the knowledge extraction process from the concept space. They converge the optimization task to the convex relations and prove that by means of optimizing the conceptual space and merging multiple concepts (or even spaces) together, one can extract new knowledge practical for the downstream task. They also provide an algebra language on how concepts are organized and interact and what it means mathematically when several concepts (or concept spaces) are combined. They put the conceptual representations in different compacts and explore the vectors’ behavior there. This is one of the ideas we are adopting in our paper, which we believe helps regularize the network. The concept spaces are encapsulated into a compact hypercube with the side 2. This is achieved due to the utilization of the $tanh$ activation, which we will review in more detail in the methodology section.

\section{Methodology}
\noindent We are going to use the transformer architecture to extract the contextual embeddings from the input text. However, the methodology is not limited to transformers and may be reused with any architecture producing some kind of embeddings. In this specific research, we are focusing on the BERT family models (and some variations such as XLNet).

\subsection{Context Attribution}
Context attribution - is a projection of a collection of contextual embeddings (vectors) or, simply, a matrix. Projection is done via a concept operator. When we train the model, we ensure that concept operators project disjoint concepts far away from similar concepts. We project the sentence in multiple context attributions and then find the similarity between the original sentence and the conceptual projections. This similarity tells how to classify the instance correctly. Note that in the actual implementation, we do not do the pairwise comparisons of the similarities or any other type of processing. Instead, we concatenate obtained projections into a single tensor and feed it to the classification layer. Thus, instead of manually defining the classification criteria, we specify it as a set of trainable parameters.

\subsection{Contextual word embeddings}

As it was already stated, the only assumption we impose on the base model is that it can create (contextual) embeddings from the input. Let $N_s$ be the sequence length, $d$ be the dimensionality of the contextualized embedding of the model. $E=[e_1, e_2, ..., e_{N_s}] \in R^{N_S \times d}$, $E_{N_s \times d} \in R^{N_S \times d}$ is the contexual embedding matrix. 

\bigskip

In our research, we assume that the BERT-like models produce this embedding. So $N_s$ would be defined in the range between 256 and 512 (as a maximum sequence length used by the BERT architecture), and $d$ would be 768 for all of the base models and 1024 for the XLNet large.

\subsection{Conceptual projections}

For each of the classes in the classification problem, we assume a single concept space operator. This concept space operator transforms (projects) the contextual embeddings to the context attribution and produces new conceptual embeddings. The obtained representation of the embeddings is also called a latent representation. This representation’s dimensionality is defined as the latent space (target space for the projection). Let $m$ be the dimensionality of the latent space. We first define the projection operator as a matrix with trainable parameters: $P_{d \times m} \in R^{d \times m}$. Thus obtained projection matrix (context attribution) $C_{N_s \times m} \in R^{N_s \times m}$. 

\bigskip

Basically, context attribution is a new representation of the embeddings in the latent space, where the transformation operator is trained during fine-tuning. However, since we want to obtain proximity of the contextualized embedding to the context attribution, we introduce previously defined $tanh$ operation as a similarity measure.

\begin{equation}
    C_{N_s \times m} = tanh(E_{N_s \times d} \times P_{d \times m})
\end{equation}

$tanh$ is applied element-wise. In that case, our conceptual matrix is a representation of how close a certain sentence is to the concept from the target context attribution (1 is very close, and -1 is from an orthogonal attribution). As an example, when we feed the word “terrible” to the context attribution that was predefined as “positive”, we expect to see -1 in the conceptual representation and 1 for a word like “great”.

\bigskip

The training objective of the model is, by taking into account multiple projections of the input embeddings, to find the projection that is most aligned with the sentence content. The similarity measure we are using is a slight modification of the cosine similarity, where normalizing the value by the vectors’ norms is replaced with the $tanh$.

\begin{equation}
tanh(x) = \frac{e^{x}-e^{-x}}{e^{x}+e^{-x}}
\end{equation}

Similarly to the original paper introducing LSTM \cite{HochSchm97} , we use $tanh$ to control the flow of the information in the network. It squashes the range, centers the values around zero, and introduces non-linearity. This has also proven to be an excellent regularization technique, which reduces the instability, improves models’ \textit{generalizability}, and improves the results. This aspect is discussed in more detail in the results section.

\begin{figure}[h]
\centering
\includegraphics*[width=0.45\textwidth]{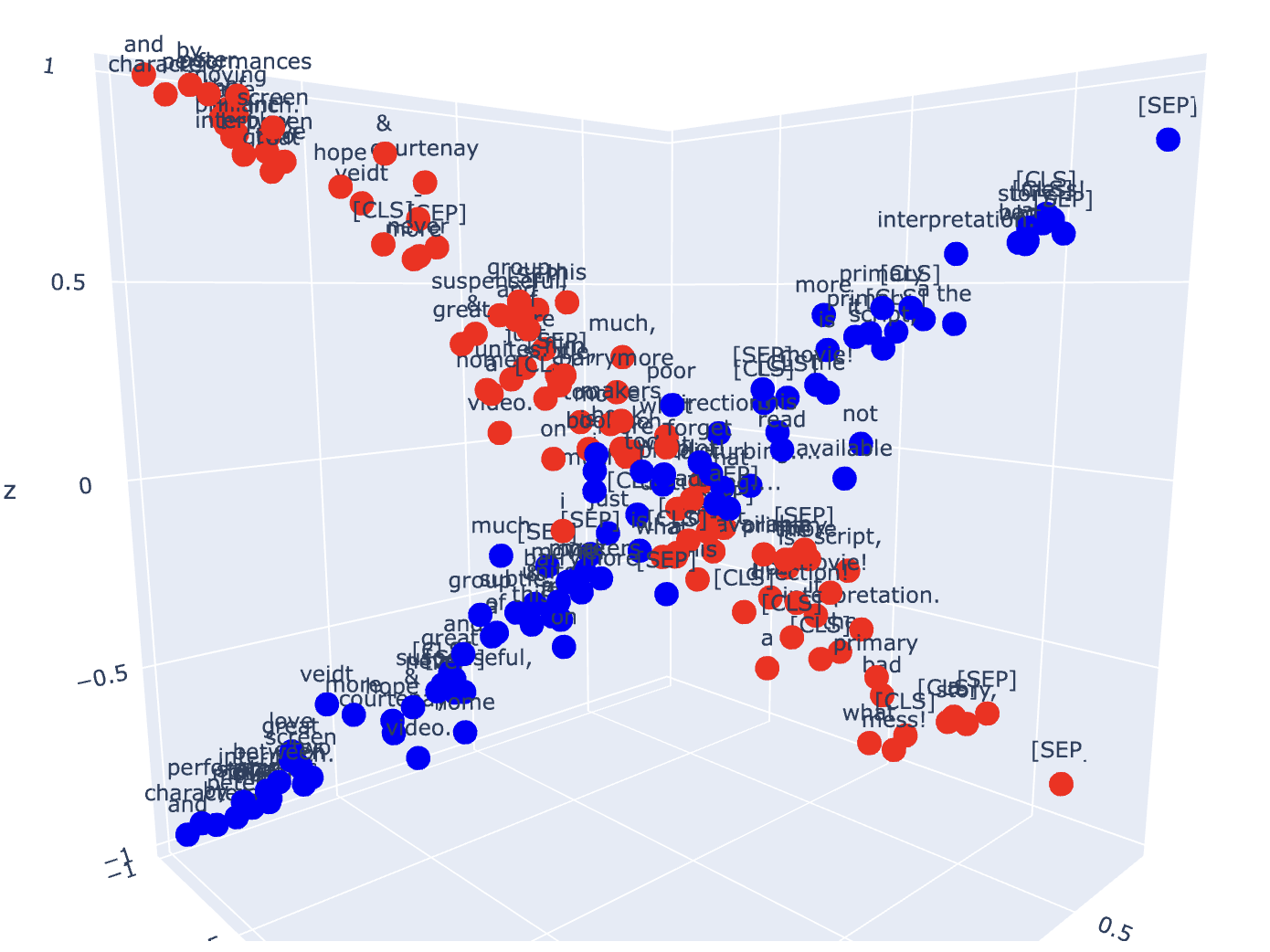}
\caption{\small 3D projection of the space embeddings for the 2-class classification. After projecting the sentence onto different concept spaces, we expect these projections to be orthogonal if the classes are completely divergent. For the case between positive and negative sentiment, we expect that positive class projection would be orthogonal to the negative class projection.}
\end{figure}

As a good side-effect of the $tanh$ we add additional non-linearity and squashing effect to the model. Thus, no additional normalization of values is required. Besides, we shrink our problem to the compact (hypercube with side 2, from -1 to 1). In Figure 1, one can find a benefit from such an approach. We can now easily interpret the outcome of the binary classification model. The visualization provided is the contextual embedding of the BERT model into 3-dimensional context attribution space for the IMDB classification task (this is done for test examples, so the model is not overfitted, and what we clearly see is the orthogonality of the negative and positive sentiment concepts).

According to our definition, every target class has a unique context attribution for itself. So for $n$ classes classification problem:

\begin{equation}
    C_{N_s \times m}^i = tanh(E_{N_s \times d} \times P_{d \times m}^i)
\end{equation}

\noindent for $i \leq n$.
Where $C^i$ and $P^i$ are namely $i$-th context attribution and concept projection operator.

\subsection{Classification}
After we have projected the embeddings to all of the context attributions, we need to perform classification. In that case, since every projection is a set of vectors (where each vector is a conceptual embedding with latent size $m$) we would find the centroid of this representation for each context attribution and then concatenate these representations. This concatenated representation is then fed to the single linear layer for classification. This basically identifies the proximity of the embedding to the corresponding context attribution. Let $k_i$ represent $i$-th context attribution centroid, and $c_{i,j}$ $j$-th conceptual embedding vector ($j$-column) of the $i$-th context attribution $C_{N_s \times m}^i$.

\begin{equation}
    k_i=\frac{1}{N_s} \cdot \sum_{j=0}^{N_s} c_{i,j}
\end{equation}

\subsection{Loss function}

The loss that we are optimizing is primarily the Cross-Entropy loss. To ensure that the conceptual embeddings don’t converge to the same embedding inside the conceptual space, we introduce an intra-space loss. This also adds additional regularization and improves \textit{generalization}. This is proved during the experiments. Controlling the weight of this loss compared to the cross entropy loss is another hyperparameter fine-tuning task. The intra-space loss is basically an inverse of the variance of the vectors inside the context attribution.

\begin{equation}
    \sigma^2 = \sum_{i=1}^{m} \frac{1}{m} \cdot (c_i - \hat{c})^2
\end{equation}

where $c_i$ is $i$-th embedding ($i$-th column of the context attribution matrix) $C_{N_s \times m}$ and $\hat{c}$ is the mean vector of conceptual embedding matrix (column-wise).

\section{Results}

We evaluate our framework using 3 base models in 3 benchmarks with 3 different datasets. 
With benchmarks, we want to measure:
\begin{itemize}
    \item Performance of the proposed Space Model
    \item Generalization property of the novel technique
    \item How does it compare with existing context attribution solutions which involve a manual process 
\end{itemize}
To achieve that, we are going to use 3 datasets, namely HateXplain, IMDB reviews sentiment dataset, and Social Media Attributions dataset of YouTube comments related to the Chennai water crisis. IMDB sentiment reviews is a semantically close dataset to the HateXplain and is reasonably easily interpretable. Such a relation is essential since we would like to test the \textit{generalizability} of the proposed technique. Besides, in the Social Media Attributions paper, the authors do manual labeling of the concepts and measure similarity with the so-called Social Media Attributions; we would like to show that our approach achieves superior performance without additional manual labeling and via a single pass.

\bigskip

The experiments are structured in a way that we have basic experiments with smaller models and simpler tasks. Additionally, we conducted experiments to compare the Space Model to the state-of-the-art model of the IMDB dataset. We also investigate and analyze various properties of the Space Model and explore some of the hyperparameters’ usage, with their respectful effect on the model performance. We explore the \textit{generalization} property of the model by cross-testing it on the unseen dataset.

\begin{table*}[t]
\small
\centering
\begin{tabular}{|l|l|c|c|c|c|c|c|}
\hline
Dataset & Model & Train Params & Accuracy & F1-score (macro) & Recall & Intra-Space weight \\
\hline
IMDB (training) & DistilBERT & 592130 & 0.7852 & 0.7819 & 0.6614 & N/A \\
\Cline{0.1pt}{2-7}
& Space Model & 197122 & 0.7917 & 0.7916 & 0.7728 & 0.001 \\
& Space Model & 197122 & \textbf{0.8322} & \textbf{0.8320} & \textbf{0.8663} & 0 \\
\hline
HateXplain (zero-shot testing) & DistilBERT & 592130 & \textbf{0.6013} & 0.4450 & 0.0869 & N/A \\
\Cline{0.1pt}{2-7}
& Space Model & 197122 & 0.5821 & \textbf{0.5187} & \textbf{0.2698} & 0.001 \\
& Space Model & 197122 & 0.5977 & 0.5040 & 0.2007 & 0 \\
\hline
\end{tabular}
\caption{\centering Comparative table of the results of the Space Model in different configurations with DistilBERT on the IMDB dataset and HateXplain dataset}
\label{tab:combined_table}
\end{table*}

\subsection{Preprocessing and settings}

\noindent BERT is a standard model that we use as a reference and a baseline. We only fully train the weights of this model once when compared with the Chennai water crisis data. For all of the other experiments, we preserve all of the \textit{generalizability} and do not spend time on training.
XLNet is the current state-of-the-art transformer for multiple benchmarks; in this specific work, we focus on the IMDB sentiment analysis dataset. By using this model and comparing the results with it, we want to prove that attaching the Space-model head to virtually any current state-of-the-art transformer would significantly boost performance.

We are not conducting any data preprocessing for either of the datasets. We use cased models for all of the expriments except for the Social Media Attributions comparison. For the space model, the key idea is the contextual embedding generation. The entity doing this in our framework is called a base model; virtually any transformer model can play this role. We use cased DistliBERT, cased BERT, and cased XLNet.

\begin{figure}[h]
\centering
\includegraphics*[width=0.45\textwidth]{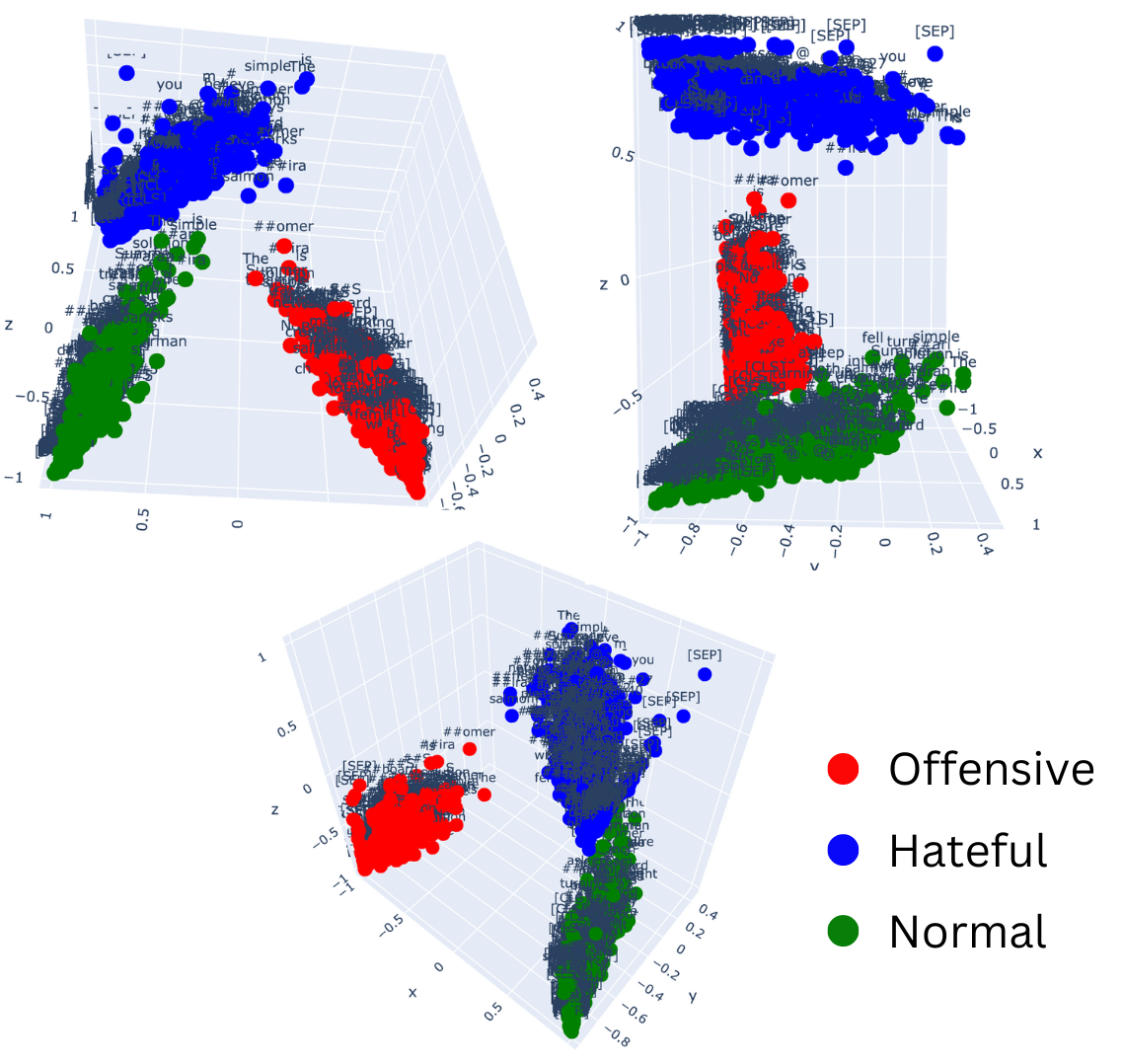}
\caption{\small 3D projection of the space embeddings for the 3-class classification (HateXplain). For the 3-class, similar to the 2-class, we expect to have 3 orthogonal projections. Here, we observe that if we review this image in multiple projections - some projections are clearly orthogonal, and some are more aligned. This is the effect that we have discussed previously, that contextual attributions might have overlapping concepts.}
\end{figure}

We use the base model configuration for all of the experiments except for the state-of-the-art establishment (12 layers for BERT and XLNet and 6 layers for DistilBERT). For the state-of-the-art performance, we trained large (24 layers) XLNet, which was used for reporting the results in the original paper. We use the Adam optimizer with a learning rate of $2 \cdot 10^{-4}$ for all experiments, except for the state-of-the-art establishment, since the original paper states that $10^{-5}$ was used to achieve the best results. Maximum sequence length and batch size is 256 for all of the basic experiments and is replaced with 512 and 4, respectively, for the XLNet large state-of-the-art results.

\bigskip

Since the original paper recommends using 32 as the batch size for the IMDB benchmark for the best results, and we could not fit that to the GPU memory, we used 8 gradient accumulation steps and adjusted the number of training steps accordingly. Even though the result does not precisely reproduce the original outcome, it is close, and the evident performance boost from the space model is transparent.

For the number of latent spaces, we use three for most experiments since this is enough to outperform significantly and is easy to visualize. As discussed previously, when we project the contextual embeddings onto the context attribution, we expect these projections to be orthogonal if the classes are different. That is what we observe in Figure 1 and Figure 2.

For the comparison with the Social Media Attributions, use the latent size of 64. For the state-of-the-art results using XLNet, we use 128 as the latent space size. We use a single Nvidia A5000 GPU for our training. Our model with various configurations may take from 30 seconds per epoch with DistilBERT to 25 minutes with XLNet large. A standard number of fine-tuning epochs is set to 5; however, for the XLNet large state-of-the-art results, we used only one epoch of training with one epoch of head fine-tuning to prevent overfitting.

\subsection{Evaluation Metrics}
Since we are evaluating the model between multiple benchmarks simultaneously, we want to adjust to both a perfectly balanced IMDB dataset and a less balanced HateXplain dataset. So, we report accuracy and f1-macro score. Our loss throughout the experiments is Cross-Entropy loss, sometimes combined with intra-space loss for better regularization. We also report the weight of the Intra-space loss in the experiments. This is usually set to a very low number to avoid dominance over the cross-entropy loss.

\subsection{Experimental Results}

\subsubsection{Fine-tuning Space Model} 
First, we ran a set of experiments on the IMDB benchmark dataset with the DistilBERT model as a base model (Table 1). We observe that the Space model is superior for both accuracy and f1-macro score. We also explore the number of trained parameters. With 3-time fewer parameters, the performance boost is already around 5\% for both metrics. We also observe that with around 128 times fewer trainable parameters, the space model performs better by almost 2\%.

\begin{table}[h]
\centering
\begin{tabular}{|l|c|c|}
\hline
Metric            & Space Model    & BERT  \\
\hline
Accuracy          & \textbf{0.5296} & 0.4485 \\
F1-score (macro)  & \textbf{0.4304} & 0.3314 \\
Precision         & \textbf{0.5431} & 0.4471 \\
Recall            & \textbf{0.5296} & 0.4485 \\
\hline
\end{tabular}
\caption{BERT HateXplain (3-class) evaluation}
\label{tab:bert_hatexplain_transposed}
\end{table}

Then, we compare these results with the experiments for a much more complex HateXplain benchmark. The choice of the datasets is non-arbitrary in that case. We want data to have the evident polarization between classes, which is aligned cross-datasets, to prove the zero-short \textit{generalization} component of our approach. BERT, DistilBERT, and XLNet are all evaluated with this benchmark against the Space Model in a 3-class and 2-class setting.

\subsubsection{Generalizability}
We take corresponding models and evaluate them on the HateXplain benchmark in a zero-shot manner (Table 1). For the sentiment analysis, negative labels are encoded as 0 and positive as 1; for the HateXplain, we encode Hateful and offensive labels as 0 and normal labels as 1. Here, we see that the Space model with intra-space loss is a top model in terms of f1-score, while the accuracy is the highest for the DistilBERT. However, accounting for the dataset imbalance, we see that DistilBERT is worse in terms of f1 by at least 6-7\% and almost 4 times worse in terms of recall.
Next, we compare the BERT model with the Space model and base BERT on the HateXplain benchmark with 3 classes. Here, we only train the classification head for BERT and contextual attribution operators for the Space model (as discussed previously). The results in Table 2 clearly show that the Space Model is superior in all of the metrics by at least 8\%.

\bigskip

We then fine-tune the classification head and the space model with XLNet and BERT base models for the same HateXplain benchmark (Table 5) and observe that for BERT, the performance gap with identical training settings and identical base model is more than 16\% on the f1-macro score. In comparison, for XLNet, this gap is around 6\%.

To further prove the effect of the performance boost using the space model, we do the full training of the XLNet, a state-of-the-art model for the IMDB benchmark (Table 3). With almost identical settings to the original paper, we obtain a 0.9386 f1-score, while training the space model with the exact same settings gives us \textbf{0.9487} (all of the other metrics are also superior for the space model, except for the recall, which is again very different with precision for the vanilla model, and very close for the space model). We observe that the space model surpasses the state-of-the-art models in the tasks and is much more tolerant to the imbalanced data. The precision-recall trade-off is evident in most of the experiments. To prove this point further, we conducted the ablation study and researched how the space model stabilizes performance during training.

\begin{table}[h]
\centering
\begin{tabular}{|l|c|c|}
\hline
Metric           & Space Model & XLNet \\
\hline
Accuracy         & \textbf{0.9488} & 0.9387 \\
F1-score (macro) & \textbf{0.9487} & 0.9386 \\
Precision        & \textbf{0.9463} & 0.9106 \\
Recall           & 0.9516 & \textbf{0.9731} \\
\hline
\end{tabular}
\caption{State-of-the-art XLNet on IMDB}
\label{tab:sota_xlnet}
\end{table}

\subsubsection{Social Media Attribution}
We compare our Space Model with the Social Media Attribution and observe 5.2\% F1-score improvement on our own reproducing experiment and almost 2\% F1-score improvement compared to the best-reported score from the original paper. The best result there was obtained on the adapted Indian BERT, while we used base uncased BERT without any adaptation, so this performance boost is not exhaustive.

\begin{table}[h]
\begin{tabular}{|l|c|c|c|c|c|}
\hline
Metric & Space-model & BERT (uncased) \\
\hline
Accuracy & \textbf{0.8309} & 0.8220 \\
F1-score (macro) & \textbf{0.8006} & 0.7484 \\
Precision & 0.7126 & \textbf{0.8876} \\
Recall & \textbf{0.7337} & 0.4674 \\
\hline
\end{tabular}
\caption{Social Media Attribution BERT-uncased}
\label{tab:sma_comparison}
\end{table}

\begin{table*}[t]
\centering
\begin{tabular}{|l|c|c|c|c|c|}
\hline
Metric              & Train Params & Accuracy   & F1-score (macro) & Precision  & Recall     \\
\hline
Space-model (XLNet) & 4622         & \textbf{0.8798} & \textbf{0.8797} & \textbf{0.8764} & \textbf{0.8824} \\
XLNet-base-cased    & 1538         & 0.8160     & 0.8156           & 0.8421     & 0.7750     \\
\hline
Space-model (BERT)  & 4622         & \textbf{0.8110} & \textbf{0.8108} & \textbf{0.8227} & \textbf{0.7899} \\
BERT-base-cased     & 1538         & 0.6588     & 0.6555           & 0.6919     & 0.5649     \\
\hline
\end{tabular}
\caption{BERT and XLNet Comparison on HateXplain Dataset (2-class)}
\label{tab:bert_vs_xlnet}
\end{table*}

\subsection{Regularization effect on fine-tuning}

\begin{figure}[h]
\centering
\includegraphics*[width=0.45\textwidth]{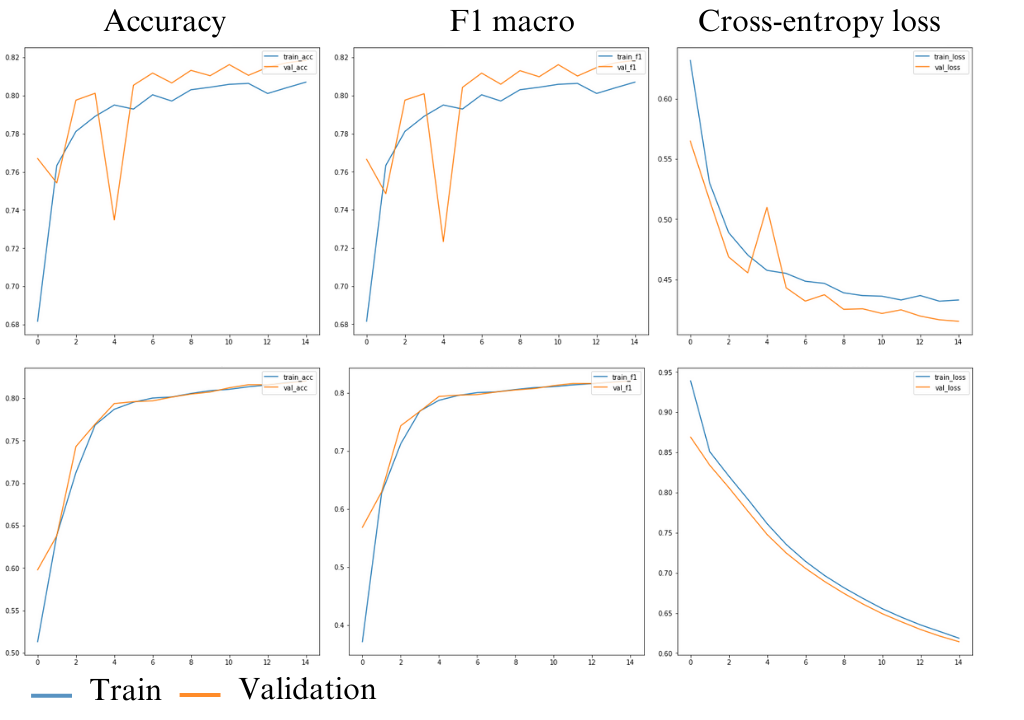}
\caption{DistilBERT (upper part) vs DistilBERT and Space Model (lower part) stabilization comparison}
\end{figure}

During the experiments, we observed that the space model has a much better ratio of recall/precision, which means that it handles imbalanced data much more efficiently. Another observation is that adding just a space model stabilizes the results during training, not allowing the performance to vary a lot between iterations. Find the visualization of the ablation study in Figure 3.
Additionally, intra-space loss adds more regularization and stabilization and ensures that the concepts in the context attribution will not converge to a single vector.

\section{Conclusion and discussion}
In conclusion, this research focused on the novel methodology towards conceptual embedding for classification with Language models. As an outcome of this research, we have conducted a set of experiments to empirically prove the efficiency of the proposed technique. We have also created the implementation of the proposed framework via PyTorch and provided an open-source GitHub repository to incivate and simplify future collaboration and exploration. We believe that the potential of this approach is yet to be discovered, and the goal of this paper was to provide some baseline ideas and understanding.

We anticipate improvements by adding more complex transformations after the conceptual projection phase. We also believe that this technique should in no way be limited to classification problems only. The formulation of the regression problem is quite straightforward but needs to be additionally researched. With that, we also expect that 1-to-1 correspondence of the context attribution to the target class is an artificial limitation that we hold in this paper for the simplicity of interpretation. However, if domain knowledge suggests that having multiple context attributions (more than the number of classes) for the task makes sense - then this should also be an option. We would also like to explore further the potential of the interpretation capabilities of the framework and how we can use it to extract knowledge from the model.

\bibliographystyle{aaai} 

\bibliography{space-model}

\end{document}